\pdfoutput=1

\documentclass[11pt]{article}

\usepackage{acl2025}
\usepackage{pythonhighlight}
\usepackage{times}
\usepackage{latexsym}

\usepackage[T1]{fontenc}

\usepackage[utf8]{inputenc}

\usepackage{microtype}
\usepackage{algpseudocode}
\usepackage{algorithm}
\usepackage{enumitem}
\usepackage{amssymb}
\useunder{\uline}{\ul}{}
\newcommand{\methodname}{\mbox{\textsc{InteGround}}\xspace}
\newcommand{\datasetname}{\mbox{\textsc{InteGround}}\xspace}

\title{\methodname: On the Evaluation of Verification and Retrieval Planning in Integrative Grounding}

\newcommand{\ust}{\ensuremath{^\spadesuit}}
\newcommand{\sjtu}{\ensuremath{^\heartsuit}}

\author{Cheng Jiayang\ust   \ \ \ \ Qianqian Zhuang\ust  \ \ \ \  Haorao Li\ust \ \ \ \ \\ \textbf{Chunkit Chan\ust} \ \ \ \ \textbf{Xin Liu\ust}  \ \ \ \ \textbf{Lin Qiu\sjtu}  \ \ \ \  \textbf{Yangqiu Song\ust} \ \ \ \ \\
\ust The Hong Kong University of Science and Technology \\
\sjtu Shanghai Jiaotong University \\
\texttt{\{jchengaj, yqsong\}@cse.ust.hk} \ \ \ \  \\ \\
}

\begin{document}
\maketitle
\begin{abstract}
Grounding large language models (LLMs) in external knowledge sources is a promising approach to ensuring faithful and accurate predictions. While existing grounding approaches work well for simple queries, many real-world information needs require synthesizing multiple pieces of evidence. We introduce "integrative grounding" — the challenge of retrieving and verifying multiple interdependent pieces of evidence to support a hypothesis query. To systematically study this problem, we repurpose data from four domains for evaluating integrative grounding capabilities. 
Our investigation reveals two critical findings: First, when verifying groundedness, while LLMs are robust to redundant evidence, they tend to rationalize using their internal knowledge when the provided grounding information is incomplete. Second, in examining retrieval planning strategies, we find that undirected planning can degrade performance through the introduction of noise, while premise abduction emerges as a promising approach due to its logical constraints. Additionally, we observe that LLMs' zero-shot self-reflection capabilities consistently enhance grounding quality. These insights provide valuable directions for developing more effective integrative grounding systems. \footnote{Our code is available at \url{https://github.com/HKUST-KnowComp/InteGround}.}

\end{abstract}

\section{Introduction}

\begin{figure*}
    \centering
    \includegraphics[width=1\textwidth]{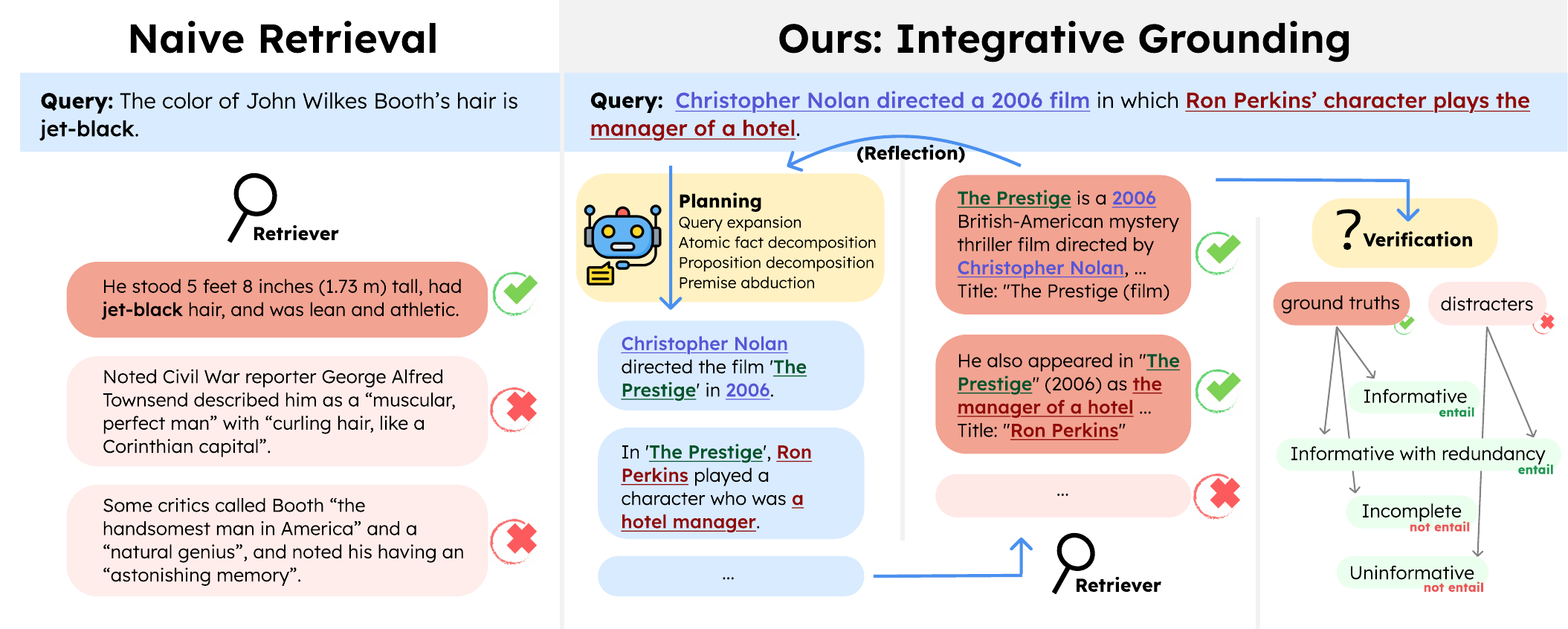}
    \caption{Overview of the integrative grounding problem.}
    \label{fig:integround}
\end{figure*}
Large language models (LLMs) are notorious for their tendency to hallucinate -- generating content that appears plausible but is factually incorrect or unsupported~\cite{ji2023survey, zhang2023siren}. 
To alleviate this issue, grounding LLMs to external knowledge sources has emerged as a promising approach.
By anchoring model outputs to verifiable information~\cite{min2023factscore, rashkin2023measuring, asai2023self}, grounding has enabled LLMs with more faithful decision-making and responsible generation.

In typical grounding setups, systems retrieve relevant documents in response to a query. 
While this approach has shown success for simple queries where a single piece of evidence suffices, many real-world information needs are inherently complex and require \textit{synthesizing multiple pieces of evidence} to form a complete answer (Figure~\ref{fig:integround}). 
For instance, answering questions about comparative analysis, multi-step reasoning~\cite{yang2018hotpotqa, trivedi2022musique}, or claims requiring evidence from different sources~\cite{min2023factscore, kamoi2023wice, dalvi2021explaining} often necessitates the integration of multiple pieces of information.
We term this problem ``\textit{integrative grounding}'': given a hypothesis query, a grounding system needs to retrieve multiple interdependent pieces of evidence to support it.

Despite its significance, the integrative grounding problem lacks systematic evaluation in current research. Existing work on Retrieval-augmented Generation (RAG) primarily focuses on end-to-end reasoning performance~\cite{yao2022react,shinn2024reflexion}, reasoning with pre-retrieved evidence~\cite{fang2024trace}, or post-generation evaluation in specific domains~\cite{trautmann2024measuring, song2024measuring}. While these approaches implicitly address aspects of integrative grounding, they offer no comprehensive analysis of the grounding problem itself. Similarly, research in automated theorem proving~\cite{dalvi2021explaining, sprague2022natural} explores complex inference chains but operates in restricted domains without robust evaluation of the broader integrative grounding challenges. While related fields like Natural Language Inference (NLI) and automated proof generation have their own evaluation paradigms, they typically operate under the assumption that sufficient evidence is provided. A critical gap remains in systematically evaluating grounding under the sub-optimal evidence conditions that are common in open-world settings. Our work addresses this gap by introducing InteGround, a novel evaluation framework designed specifically for \textit{integrative grounding}. Unlike standard benchmarks, InteGround systematically tests a model’s behavior across four distinct evidence scenarios: complete, redundant, incomplete, and uninformative. This allows us to rigorously analyze critical failure modes like rationalization—where models invent facts to fill evidence gaps—that are often missed by traditional evaluations. We argue that robust integrative grounding is a crucial prerequisite for building truly reliable and faithful generative systems.


In this work, we focus on two critical aspects for evaluating integrative grounding. First, we investigate whether models can effectively \textit{verify groundedness} by determining if multiple evidence pieces collectively support a query hypothesis (\textbf{RQ1}). Second, recognizing the challenges posed by reasoning dependencies among evidence, we examine \textit{effective planning strategies} for LLMs to reformulate search queries and guide the retrieval process (\textbf{RQ2}).

To address these questions, we first construct an evaluation dataset from four domains to evaluate a system's ability to integrate multiple pieces of evidence. 
For RQ1, we conduct experiments on \textit{groundedness verification}. Our experiments reveal that although LLMs are robust to redundant or distracting evidence, they exhibit a strong tendency to compensate for incomplete information by drawing on internal knowledge instead of strictly adhering to the retrieved content.

For RQ2, we present a systematic investigation of different \textit{planning strategies for retrieval}, including decomposition-based, query expansion-based, and premise abduction-based approaches. Our findings reveal that planning does not universally improve retrieval performance. In fact, undirected planning can degrade performance by introducing noise, while decomposition-based planning shows limited improvement due to its conservative nature. 
Notably, we identify premise abduction as a particularly promising approach that shows consistent improvements and generalizes well to other datasets. This is likely due to its strong logical constraints, which encourage a directed expansion of the search space.
Furthermore, we demonstrate that leveraging zero-shot self-reflection consistently enhances grounding quality across all planning strategies, highlighting the value of iterative refinement.

\section{Preliminaries}
\label{sec:def-preliminaries}
\noindent \textbf{Proposition. } A proposition is a statement that has a truth value. For instance ``The sky is green.'' is a proposition since it can be verified as true or false, while ``Look up at the sky'' is not a proposition.
We use $p$ to denote general propositions. For convenience, we use $\phi$ to denote a proposition that serves as hypothesis. 
In this paper, the queries are all propositions.

\noindent \textbf{Knowledge Base (KB).} A knowledge base $\mathcal{K}=\{p_i\}_{i=1, 2, ..., K}$ is a set of consistent propositions. 

\noindent \textbf{Asking a KB.} The \texttt{Ask$_{\mathcal{K}}$($p$)} operation queries KB $\mathcal{K}$ about proposition $p$, which returns three possible responses: \texttt{Entailment} ($\mathcal{K}\models p$), \texttt{Contradiction} ($\mathcal{K}\models \neg p$) , and \texttt{Contingent} (neither of the above). 
These two responses are considered as \textit{informative} as they indicate that $\mathcal{K}$ contains related knowledge about $p$. 
Here, the operator $\models$ tells whether $p$ follows logically from the premises in $\mathcal{K}$, where it is impossible for the premises to be true and $p$ to be false.
For example,  $\phi$=``Socrates is mortal.'' \textit{deductively} follows from $\mathcal{K}$=\{``All men are mortal.'', ``Socrates is a man.''\},  i.e. $\mathcal{K}\models \phi$, (\texttt{Ask}$_{\mathcal{K}}(p)$ = \texttt{Entailment}).

\noindent \textbf{Grounding.} Given a hypothesis proposition $\phi$ and a knowledge base $\mathcal{K}$, the task of \textit{grounding} is to find a subset of consistent premises $\Sigma \subseteq \mathcal{K}$ through planning and retrieval, such that $\Sigma$ is informative enough to ground the query hypothesis (i.e., $\Sigma\models p$ or $\Sigma \models \neg p$).
In practice, $\Sigma$ is obtained through the top-n retrieval results.

\section{Constructing evaluation data}

\begin{table*}[ht!]
    \resizebox{\textwidth}{!}{%
        \begin{tabular}{p{4cm}p{14cm}p{2cm}}
        \toprule
        \multicolumn{1}{c}{\textbf{Hypothesis $\phi$}} & \multicolumn{1}{c}{\textbf{Candidate Evidence KB $\mathcal{K}$}} & \multicolumn{1}{c}{\textbf{Domain}} \\
        \midrule

        {\color{purple}Northern hemisphere} will have {\color{blue}the most sunlight in summer}.
        & 
        {\color{purple}The northern hemisphere is a kind of hemisphere of earth.}  //  {\color{purple}A hemisphere of earth is a kind of place.}  //  {\color{blue}If a place is in summer, then it will have the most sunlight.} 

        {\color{gray}If an object/something is in the sunlight then that object/that something will absorb solar energy.}  //  {\color{gray}Daylight is when the sun shines on a location.}  //  {\color{gray}The northern hemisphere is a kind of hemisphere of earth.}  //  {\color{gray}...}    
        & 
        {\color{white}ab}Ent-Bank \\
        \hline

        {\color{purple}Salih won the election} with {\color{blue}219 votes to 22}.
        & 
        {\color{purple}(meta data) TITLE: Iraq: Parliament elects Barham Salih as new president | News | Al Jazeera}  //  {\color{blue}Salih routed his main rival, Fuad Hussein, with 219 votes to 22.}  
        
        {\color{gray}The Kurdish moderate politician has named veteran Shia politician Adel Abdul Mahdi as prime minister-designate.}  //  {\color{gray}Salih is a former deputy prime minister of the Iraqi federal government [Reuters]}  //  {\color{gray}...}
        & 
        {\color{white}abc}WiCE\\
        \hline

        {\color{purple}Christopher Nolan directed a 2006 film} in which {\color{blue}Ron Perkins' character plays the manager of a hotel}.
        & 
        {\color{purple}The Prestige is a 2006 British-American mystery thriller film directed by Christopher Nolan, from a screenplay ... from Christopher Priest's 1995 novel of the same name. Title: "The Prestige (film)"}  //  {\color{blue}Ron Perkins is an American actor who has been active since the early 1960s. Title: "Ron Perkins"}  //  {\color{blue}He also appeared in "The Prestige" (2006) as the manager of a hotel visited by Hugh Jackman's character in Colorado Springs, as well as ...} 
        
        {\color{gray}Inland Empire is an internationally co-produced 2006 film written and directed by David Lynch. Title: "Inland Empire (film)"}  //  {\color{gray}The film is a co-production of France, Poland and the United States. Title: "Inland Empire (film)"}  //  {\color{gray}...}
        & 
        {\color{white}ab}HotpotQA\\
        \hline

        {\color{purple}The maker of the Acura Legend}, {\color{purple}the manufacturer of the Scion xB}, and {\color{purple}Nissan}  {\color{blue}opened US assembly plants in 1981}.
        & 
        {\color{purple}The Acura Legend is a mid-size luxury/executive car manufactured by Honda. It was sold ...}  //  {\color{purple}The Scion xB is a compact car (subcompact car in its first generation) made by Toyota for the United States market and sold ...}  //  {\color{blue}... A decade after the 1973 oil crisis, Honda, Toyota and Nissan, affected by the 1981 voluntary export restraints, opened US assembly plants and established their luxury divisions (Acura, Lexus and Infiniti, respectively) to ...}
        
        {\color{gray}The Nissan Rogue is a compact crossover SUV produced by the Japanese automaker Nissan. It made its debut in October 2007 for the 2008 model year...}  //  {\color{gray}The Acura EL is a subcompact executive car that was built at Honda\'s Alliston, Ontario, plant...}  //  {\color{gray}... }
        & 
        {\color{white}ab}MuSiQue\\
        \bottomrule
    \end{tabular}
    }

    \caption{Example instances in the evaluation data. Related information in the Hypothesis and Candidate Evidence columns is color-coded for easier identification, using {\color{purple}purple} and {\color{blue}blue}. Distracting evidence are marked with {\color{gray} gray}. Due to large number of candidate instances, only part of distracting evidence are shown and the rest are left out. ``Ent-Bank'' is short for ``EntailmentBank''.} 
    \label{tab:dataset_examples}
\end{table*}

Our evaluation is based on data repurposed from two tasks where integrative grounding is required: multiple premise entailment~\cite{lai2017natural, dalvi2021explaining, kamoi2023wice} and multi-hop question answering (QA)~\cite{yang2018hotpotqa, trivedi2022musique}.

\subsection{Evaluation formulation}

In the evaluation data (examples shown in Table~\ref{tab:dataset_examples}), each hypothesis $\phi$ is accompanied by a set of ground-truth evidence $\Sigma^{gt}=\{p^{gt}_1, p^{gt}_2, \cdots\} \subseteq \mathcal{K}$ and a larger set of candidate evidence $\mathcal{K}=\Sigma^{gt}\cup\Sigma^{distr}$ that includes both the ground-truth and additional distracting facts.

For groundedness verification evaluation (Section~\ref{sec:groundedness_verification}), we test verification models' ability to accurately classify whether a target query $\phi$ is grounded by retrieval results $\Sigma$.
For retrieval planning evaluation (Section~\ref{sec:retrieval_planning}), we assess how effectively different planning strategies retrieve relevant evidence. Given a hypothesis $\phi$ and a candidate evidence set $\mathcal{K}$, an integrative grounding system retrieves related evidence through query planning (as shown in Figure~\ref{fig:integround}).



\subsection{Data composition}
We construct our evaluation dataset from four data sources, totaling 1,625 items.
Dataset examples are shown in Table~\ref{tab:dataset_examples}.

\paragraph{Multi-premise Entailment.}

\begin{itemize}[leftmargin=*]
    \item \textsc{EntailmentBank}~\cite{dalvi2021explaining} contains hypotheses and corresponding multi-step entailment tree annotations from the science facts in WorldTree~\cite{xie2020worldtree}. 
We adapt the test set under task 2 setting\footnote{\url{https://allenai.org/data/entailmentbank}. We use the split under \url{v3_May6_2022/entailment_trees_emnlp2021_data_v3/dataset/task_2/}.} for our use, where the leaf nodes of entailment trees are kept as ground-truth evidence.
Since task 2 already provides a set of (hard) distractors, we follow their setting and treat all candidate evidence for a given hypothesis as its corresponding KB, $\mathcal{K}$.
    \item \textsc{WiCE}~\cite{kamoi2023wice} is a fine-grained textual entailment dataset linking natural claims and Wikipedia evidence.
We consider both the claim and sub-claim level annotations, where for each claim several distinct groups of ground-truth evidence are annotated. We treat the evidence set for each claim as the KB $\mathcal{K}$, filtering out instances with more than 200 evidence items.
Because WiCE often provides multiple valid evidence sets for a single hypothesis, we filter out instances where there is any single answer evidence (i.e., \#GT == 1).
Furthermore, to simplify evaluation, we select only the first group of ground-truth evidence ids as the ground truth.
We only retain instances with the \texttt{supported} label.
\end{itemize}

\paragraph{Multi-hop QA.}
In the literature, a question-answer pair can be seen as a hypothesis~\cite{dalvi2021explaining}.
For multi-hop QA datasets, we prompt an LLM to transform each question and its corresponding answer into an equivalent hypothesis. 
As test sets are not always publicly available, we sample 500 instances from the validation sets of these datasets, specifically selecting those that require at least 3 pieces of evidence to answer.

\begin{itemize}[leftmargin=*]
    \item \textsc{HotpotQA}~\cite{yang2018hotpotqa} is a dataset of question-answer pairs derived from Wikipedia, designed to evaluate complex reasoning and explanation generation. The questions in this dataset necessitate finding and reasoning over multiple supporting documents to formulate answers. To create evidence pieces, we append the document titles to the end of corresponding sentences. The evidences with titles that appear in the supporting facts are considered ground truths.
    \item \textsc{MuSiQue}~\cite{trivedi2022musique} is created by composing questions from single-hop datasets. To create pieces of evidence, we concatenate the titles to the end of the corresponding paragraph texts to preserve context information. We treat the evidences corresponding to paragraph support indices as ground-truth evidences, and all others as distracting evidence.
\end{itemize}






\subsection{Overview}

\begin{table}
    \centering
\resizebox{0.9\linewidth}{!}{
\begin{tabular}{l|lll}
\toprule
               & \multicolumn{1}{c}{Items} & \multicolumn{1}{c}{$|\Sigma^{gt}|$} & \multicolumn{1}{c}{\cellcolor[HTML]{FFFFFF}$|\mathcal{K}|$} \\ \hline
EntailmentBank & 340                       & 4.5±2.4                             & 25.0±0.0                                                    \\ 
WiCE           & 285                       & 2.8±0.9                             & 85.2±43.4                                                   \\ 
HotpotQA       & 500                       & 3.4±0.6                             & 42.7±10.9                                                   \\ 
MuSiQue        & 500                       & 3.4±0.5                             & 20.0±0.1                                                    \\  \bottomrule
\end{tabular}
    }
    \caption{Statistics of \datasetname. $|\Sigma^{gt}|$ is the number of ground-truth snippets. $|\mathcal{K}|$ is the number of all candidate snippets.}
    \label{tab:data-statistics}
\end{table}

Table~\ref{tab:data-statistics} presents a comprehensive overview of all four data sources, detailing the number of items, and the means and standard deviations of ground-truth evidence ($|\Sigma^{gt}|$) and candidate evidence ($|\mathcal{K}|$) numbers.

As shown in Table~\ref{tab:dataset_examples} and~\ref{tab:data-statistics}, the evaluation data represent diverse domains and complexity levels. EntailmentBank, WiCE, HotpotQA, and MuSiQue each present distinct challenges—from logical deduction to information synthesis from news content to connecting facts across multiple sources. This diversity allows for comprehensive evaluation of integrative grounding capabilities across different contexts.

\section{Groundedness verification}
\label{sec:groundedness_verification}
\begin{table*}[ht!]
\centering
\resizebox{\textwidth}{!}{
\begin{tabular}{lrrrrrrrrrrrrrrrr}
\toprule
\multicolumn{1}{l|}{}                    & \multicolumn{4}{c|}{\textbf{EntailmentBank}}                                                                                                   & \multicolumn{4}{c|}{\textbf{WiCE}}                                                                                                              & \multicolumn{4}{c|}{\textbf{HotpotQA}}                                                                                                         & \multicolumn{4}{c}{\textbf{MuSiQue}}                                                                                      \\
\multicolumn{1}{l|}{\multirow{-2}{*}{}}  & \multicolumn{1}{c}{Info.}    & \multicolumn{1}{c}{Redun.}   & \multicolumn{1}{c}{Inc.}     & \multicolumn{1}{c|}{Uninfo.}                      & \multicolumn{1}{c}{Info.}    & \multicolumn{1}{c}{Redun.}   & \multicolumn{1}{c}{Inc.}     & \multicolumn{1}{c|}{Uninfo.}                       & \multicolumn{1}{c}{Info.}    & \multicolumn{1}{c}{Redun.}   & \multicolumn{1}{c}{Inc.}     & \multicolumn{1}{c|}{Uninfo.}                      & \multicolumn{1}{c}{Info.}    & \multicolumn{1}{c}{Redun.}   & \multicolumn{1}{c}{Inc.}     & \multicolumn{1}{c}{Uninfo.}  \\ \hline
\multicolumn{17}{l}{\cellcolor[HTML]{D9D9D9}\textit{NLI models}}                                                                                                                                                                                                                                                                                                                                                                                                                                                                                                                                                         \\
\multicolumn{1}{l|}{NLI-xxlarge}         & \cellcolor[HTML]{86CEAB}76.8 & \cellcolor[HTML]{88CFAC}75.9 & \cellcolor[HTML]{79C9A2}83.5 & \multicolumn{1}{r|}{\cellcolor[HTML]{65C194}93.2} & \cellcolor[HTML]{C5E8D7}45.6 & \cellcolor[HTML]{CCEBDC}42.1 & \cellcolor[HTML]{5DBE8E}97.5 & \multicolumn{1}{r|}{\cellcolor[HTML]{58BC8B}99.6}  & \cellcolor[HTML]{7BCAA4}82.2 & \cellcolor[HTML]{87CFAC}76.4 & \cellcolor[HTML]{65C194}93.2 & \multicolumn{1}{r|}{\cellcolor[HTML]{60BF90}96.0} & \cellcolor[HTML]{C9EADA}43.6 & \cellcolor[HTML]{ECF7F2}26.6 & \cellcolor[HTML]{5CBD8D}98.0 & \cellcolor[HTML]{5ABD8C}98.6 \\
\multicolumn{1}{l|}{NLI-xlarge}          & \cellcolor[HTML]{72C69D}87.1 & \cellcolor[HTML]{72C69D}87.1 & \cellcolor[HTML]{81CCA7}79.4 & \multicolumn{1}{r|}{\cellcolor[HTML]{6FC59B}88.2} & \cellcolor[HTML]{B3E1CA}54.7 & \cellcolor[HTML]{B9E3CE}51.9 & \cellcolor[HTML]{5FBF90}96.1 & \multicolumn{1}{r|}{\cellcolor[HTML]{57BB8A}100.0} & \cellcolor[HTML]{7BCAA4}82.2 & \cellcolor[HTML]{81CCA8}79.2 & \cellcolor[HTML]{6BC498}90.2 & \multicolumn{1}{r|}{\cellcolor[HTML]{62C092}94.8} & \cellcolor[HTML]{C1E6D4}47.6 & \cellcolor[HTML]{BDE5D1}49.8 & \cellcolor[HTML]{62C092}94.8 & \cellcolor[HTML]{66C295}92.6 \\
\multicolumn{17}{l}{\cellcolor[HTML]{D9D9D9}\textit{LLMs}}                                                                                                                                                                                                                                                                                                                                                                                                                                                                                                                                                               \\
\multicolumn{1}{l|}{Llama3.1 8B Instr.}  & \cellcolor[HTML]{60BF90}95.9 & \cellcolor[HTML]{60BF90}95.9 & \cellcolor[HTML]{FBFEFD}18.8 & \multicolumn{1}{r|}{\cellcolor[HTML]{DDF1E7}34.1} & \cellcolor[HTML]{71C69C}87.4 & \cellcolor[HTML]{77C8A1}84.2 & \cellcolor[HTML]{E8F6EF}28.4 & \multicolumn{1}{r|}{\cellcolor[HTML]{96D5B6}69.1}  & \cellcolor[HTML]{7ACAA3}82.8 & \cellcolor[HTML]{7CCAA4}82.0 & \cellcolor[HTML]{B2E0CA}55.0 & \multicolumn{1}{r|}{\cellcolor[HTML]{7BCAA3}82.6} & \cellcolor[HTML]{D1EDDF}40.0 & \cellcolor[HTML]{DAF0E5}35.6 & \cellcolor[HTML]{78C9A1}84.0 & \cellcolor[HTML]{6AC397}91.0 \\
\multicolumn{1}{l|}{Llama3.1 70B Instr.} & \cellcolor[HTML]{5DBE8E}97.4 & \cellcolor[HTML]{5CBD8D}97.9 & \cellcolor[HTML]{ECF8F2}26.5 & \multicolumn{1}{r|}{\cellcolor[HTML]{D0ECDE}40.3} & \cellcolor[HTML]{79C9A2}83.2 & \cellcolor[HTML]{7BCAA3}82.5 & \cellcolor[HTML]{BEE5D2}49.1 & \multicolumn{1}{r|}{\cellcolor[HTML]{73C79E}86.3}  & \cellcolor[HTML]{70C59C}87.8 & \cellcolor[HTML]{75C79F}85.4 & \cellcolor[HTML]{96D5B6}69.2 & \multicolumn{1}{r|}{\cellcolor[HTML]{6CC499}89.8} & \cellcolor[HTML]{ABDDC5}58.6 & \cellcolor[HTML]{BAE3CF}51.4 & \cellcolor[HTML]{77C8A0}84.6 & \cellcolor[HTML]{63C093}94.2 \\
\multicolumn{1}{l|}{Claude3 Haiku}       & \cellcolor[HTML]{5EBE8F}96.8 & \cellcolor[HTML]{5DBE8F}97.1 & \cellcolor[HTML]{F8FDFB}20.3 & \multicolumn{1}{r|}{\cellcolor[HTML]{E8F6EF}28.5} & \cellcolor[HTML]{72C69D}86.7 & \cellcolor[HTML]{79C9A2}83.2 & \cellcolor[HTML]{A6DBC1}61.1 & \multicolumn{1}{r|}{\cellcolor[HTML]{73C79E}86.3}  & \cellcolor[HTML]{7ECBA6}80.8 & \cellcolor[HTML]{82CDA8}78.8 & \cellcolor[HTML]{92D3B4}70.8 & \multicolumn{1}{r|}{\cellcolor[HTML]{68C296}91.8} & \cellcolor[HTML]{DAF0E5}35.6 & \cellcolor[HTML]{E7F6EF}28.8 & \cellcolor[HTML]{6DC499}89.6 & \cellcolor[HTML]{5EBE8F}96.8 \\
\multicolumn{1}{l|}{Claude3 Sonnet}      & \cellcolor[HTML]{58BC8B}99.7 & \cellcolor[HTML]{59BC8B}99.4 & \cellcolor[HTML]{FFFFFF}16.8 & \multicolumn{1}{r|}{\cellcolor[HTML]{CAEADA}43.5} & \cellcolor[HTML]{79C9A2}83.2 & \cellcolor[HTML]{77C8A0}84.6 & \cellcolor[HTML]{9AD6B9}67.0 & \multicolumn{1}{r|}{\cellcolor[HTML]{68C296}91.6}  & \cellcolor[HTML]{73C69D}86.6 & \cellcolor[HTML]{78C9A1}84.0 & \cellcolor[HTML]{9DD8BB}65.4 & \multicolumn{1}{r|}{\cellcolor[HTML]{70C59B}88.0} & \cellcolor[HTML]{C1E6D4}47.6 & \cellcolor[HTML]{CCEBDC}42.4 & \cellcolor[HTML]{73C69D}86.6 & \cellcolor[HTML]{62C092}94.6 \\
\multicolumn{1}{l|}{Claude3.5 Sonnet}    & \cellcolor[HTML]{75C89F}85.3 & \cellcolor[HTML]{78C8A1}84.1 & \cellcolor[HTML]{98D6B8}67.9 & \multicolumn{1}{r|}{\cellcolor[HTML]{6EC59A}88.8} & \cellcolor[HTML]{C8E9D9}44.2 & \cellcolor[HTML]{D3EDE0}38.9 & \cellcolor[HTML]{5DBE8E}97.2 & \multicolumn{1}{r|}{\cellcolor[HTML]{59BC8B}99.3}  & \cellcolor[HTML]{9AD6B9}67.0 & \cellcolor[HTML]{99D6B8}67.6 & \cellcolor[HTML]{62C092}95.0 & \multicolumn{1}{r|}{\cellcolor[HTML]{5CBD8E}97.8} & \cellcolor[HTML]{E2F4EB}31.4 & \cellcolor[HTML]{EBF7F1}26.8 & \cellcolor[HTML]{5BBD8D}98.4 & \cellcolor[HTML]{5BBD8D}98.2 \\
\multicolumn{1}{l|}{GPT-4o}              & \cellcolor[HTML]{5DBE8E}97.4 & \cellcolor[HTML]{5DBE8F}97.1 & \cellcolor[HTML]{DCF1E7}34.4 & \multicolumn{1}{r|}{\cellcolor[HTML]{B4E1CB}54.4} & \cellcolor[HTML]{97D5B7}68.4 & \cellcolor[HTML]{A1D9BE}63.5 & \cellcolor[HTML]{75C89F}85.3 & \multicolumn{1}{r|}{\cellcolor[HTML]{5EBE8F}96.8}  & \cellcolor[HTML]{7BCAA4}82.2 & \cellcolor[HTML]{75C79F}85.4 & \cellcolor[HTML]{8AD0AE}74.8 & \multicolumn{1}{r|}{\cellcolor[HTML]{67C295}92.2} & \cellcolor[HTML]{C0E6D4}48.0 & \cellcolor[HTML]{B9E3CF}51.6 & \cellcolor[HTML]{77C8A1}84.2 & \cellcolor[HTML]{61BF91}95.2 \\  \bottomrule
\end{tabular}
}
\caption{Accuracy (\%) of verification methods on different types (\underline{Info}rmative, Informative with \underline{Redun}dancy, \underline{Inc}omplete, \underline{Uninfo}rmative) of items across datasets. Since it is a two-way classification, chance accuracy is 50\%.}
\label{tab:verification-results}
\end{table*}
\begin{table*}[ht!]
\centering
\resizebox{0.8\textwidth}{!}{
\begin{tabular}{lrrrrrrrrrrrr}
\toprule
\multicolumn{1}{l|}{}                    & \multicolumn{3}{c|}{\textbf{EntailmentBank}}                                                                    & \multicolumn{3}{c|}{\textbf{WiCE}}                                                                              & \multicolumn{3}{c|}{\textbf{HotpotQA}}                                                                          & \multicolumn{3}{c}{\textbf{MuSiQue}}                                                       \\
\multicolumn{1}{l|}{\multirow{-2}{*}{}}  & \multicolumn{1}{c}{P}        & \multicolumn{1}{c}{R}        & \multicolumn{1}{c|}{F1}                           & \multicolumn{1}{c}{P}        & \multicolumn{1}{c}{R}        & \multicolumn{1}{c|}{F1}                           & \multicolumn{1}{c}{P}        & \multicolumn{1}{c}{R}        & \multicolumn{1}{c|}{F1}                           & \multicolumn{1}{c}{P}        & \multicolumn{1}{c}{R}        & \multicolumn{1}{c}{F1}       \\ \hline
\multicolumn{13}{l}{\cellcolor[HTML]{D9D9D9}\textit{NLI models}}                                                                                                                                                                                                                                                                                                                                                                                                                            \\
\multicolumn{1}{l|}{NLI-xxlarge}         & \cellcolor[HTML]{76C8A0}86.8 & \cellcolor[HTML]{8FD2B1}76.3 & \multicolumn{1}{r|}{\cellcolor[HTML]{83CDA9}81.2} & \cellcolor[HTML]{5EBE8F}96.9 & \cellcolor[HTML]{DCF1E7}43.9 & \multicolumn{1}{r|}{\cellcolor[HTML]{B5E1CC}60.4} & \cellcolor[HTML]{66C194}93.6 & \cellcolor[HTML]{88CFAC}79.3 & \multicolumn{1}{r|}{\cellcolor[HTML]{78C9A1}85.9} & \cellcolor[HTML]{62C091}95.4 & \cellcolor[HTML]{F1FAF6}35.1 & \cellcolor[HTML]{CBEADB}51.3 \\
\multicolumn{1}{l|}{NLI-xlarge}          & \cellcolor[HTML]{7CCAA4}84.3 & \cellcolor[HTML]{75C89F}87.1 & \multicolumn{1}{r|}{\cellcolor[HTML]{79C9A2}85.7} & \cellcolor[HTML]{5FBE90}96.5 & \cellcolor[HTML]{C6E8D7}53.3 & \multicolumn{1}{r|}{\cellcolor[HTML]{A1D9BE}68.7} & \cellcolor[HTML]{6BC398}91.5 & \cellcolor[HTML]{85CEAA}80.7 & \multicolumn{1}{r|}{\cellcolor[HTML]{78C9A1}85.8} & \cellcolor[HTML]{72C69D}88.5 & \cellcolor[HTML]{D1EDDF}48.7 & \cellcolor[HTML]{AFDFC8}62.8 \\
\multicolumn{13}{l}{\cellcolor[HTML]{D9D9D9}\textit{LLMs}}                                                                                                                                                                                                                                                                                                                                                                                                                                  \\
\multicolumn{1}{l|}{Llama3.1 8B Instr.}  & \cellcolor[HTML]{BEE5D2}56.7 & \cellcolor[HTML]{60BF90}96.0 & \multicolumn{1}{r|}{\cellcolor[HTML]{9BD7B9}71.3} & \cellcolor[HTML]{ACDEC5}64.2 & \cellcolor[HTML]{74C79E}87.6 & \multicolumn{1}{r|}{\cellcolor[HTML]{94D4B5}74.1} & \cellcolor[HTML]{95D5B6}73.6 & \cellcolor[HTML]{80CCA7}82.5 & \multicolumn{1}{r|}{\cellcolor[HTML]{8BD1AF}77.8} & \cellcolor[HTML]{8CD1AF}77.5 & \cellcolor[HTML]{EAF7F1}38.0 & \cellcolor[HTML]{CBEADB}51.0 \\
\multicolumn{1}{l|}{Llama3.1 70B Instr.} & \cellcolor[HTML]{B7E2CD}59.4 & \cellcolor[HTML]{5CBD8E}97.6 & \multicolumn{1}{r|}{\cellcolor[HTML]{95D4B5}73.9} & \cellcolor[HTML]{99D6B8}72.0 & \cellcolor[HTML]{80CCA6}82.8 & \multicolumn{1}{r|}{\cellcolor[HTML]{8DD1B0}77.0} & \cellcolor[HTML]{84CEAA}80.9 & \cellcolor[HTML]{76C8A0}86.6 & \multicolumn{1}{r|}{\cellcolor[HTML]{7ECBA5}83.6} & \cellcolor[HTML]{7DCBA5}83.8 & \cellcolor[HTML]{C2E7D5}55.0 & \cellcolor[HTML]{A7DCC2}66.4 \\
\multicolumn{1}{l|}{Claude3 Haiku}       & \cellcolor[HTML]{BFE5D3}56.2 & \cellcolor[HTML]{5EBE8F}96.9 & \multicolumn{1}{r|}{\cellcolor[HTML]{9BD7BA}71.1} & \cellcolor[HTML]{8AD0AE}78.3 & \cellcolor[HTML]{7BCAA3}84.9 & \multicolumn{1}{r|}{\cellcolor[HTML]{83CDA9}81.5} & \cellcolor[HTML]{82CDA8}81.8 & \cellcolor[HTML]{87CFAB}79.8 & \multicolumn{1}{r|}{\cellcolor[HTML]{84CEAA}80.8} & \cellcolor[HTML]{7ACAA3}85.0 & \cellcolor[HTML]{F8FDFA}32.2 & \cellcolor[HTML]{D6EFE2}46.7 \\
\multicolumn{1}{l|}{Claude3 Sonnet}      & \cellcolor[HTML]{B9E3CE}58.8 & \cellcolor[HTML]{57BB8A}99.6 & \multicolumn{1}{r|}{\cellcolor[HTML]{95D4B5}73.9} & \cellcolor[HTML]{86CEAB}80.2 & \cellcolor[HTML]{7DCBA5}83.9 & \multicolumn{1}{r|}{\cellcolor[HTML]{81CCA8}82.0} & \cellcolor[HTML]{8AD0AE}78.5 & \cellcolor[HTML]{7AC9A2}85.3 & \multicolumn{1}{r|}{\cellcolor[HTML]{82CDA8}81.8} & \cellcolor[HTML]{80CCA7}82.7 & \cellcolor[HTML]{DAF0E5}45.0 & \cellcolor[HTML]{BAE3CF}58.3 \\
\multicolumn{1}{l|}{Claude3.5 Sonnet}    & \cellcolor[HTML]{87CFAC}79.7 & \cellcolor[HTML]{7BCAA3}84.7 & \multicolumn{1}{r|}{\cellcolor[HTML]{81CCA8}82.1} & \cellcolor[HTML]{60BF90}96.0 & \cellcolor[HTML]{E2F3EB}41.6 & \multicolumn{1}{r|}{\cellcolor[HTML]{BBE4D0}58.0} & \cellcolor[HTML]{63C092}94.9 & \cellcolor[HTML]{A4DBC0}67.3 & \multicolumn{1}{r|}{\cellcolor[HTML]{89D0AD}78.8} & \cellcolor[HTML]{64C093}94.5 & \cellcolor[HTML]{FFFFFF}29.1 & \cellcolor[HTML]{DBF1E6}44.5 \\
\multicolumn{1}{l|}{GPT-4o}              & \cellcolor[HTML]{ADDEC6}63.6 & \cellcolor[HTML]{5DBE8E}97.2 & \multicolumn{1}{r|}{\cellcolor[HTML]{8ED1B0}76.9} & \cellcolor[HTML]{73C79E}88.1 & \cellcolor[HTML]{A8DCC2}66.0 & \multicolumn{1}{r|}{\cellcolor[HTML]{91D3B3}75.4} & \cellcolor[HTML]{7ECBA5}83.5 & \cellcolor[HTML]{7DCBA5}83.8 & \multicolumn{1}{r|}{\cellcolor[HTML]{7DCBA5}83.7} & \cellcolor[HTML]{7FCCA6}82.9 & \cellcolor[HTML]{CEECDD}49.8 & \cellcolor[HTML]{B1E0C9}62.2 \\
\bottomrule
\end{tabular}
}
\caption{Classification performance (\%) of verification methods. }
\label{tab:verification-prf1-results}
\end{table*}

In this section, we aim to investigate whether models are capable of doing groundedness verification (RQ1).

\subsection{Evaluation setup}

\paragraph{Evaluation set creation}
Given candidate evidence set $\hat \Sigma_t$, we aim to test verification methods' classification performance on whether $\hat\Sigma_t$ provides informative enough clues for hypothesis $\phi$.

We randomly sample from the retrieval datasets to construct such set. 
Specifically,  we consider four cases: 
(1) informative ($\hat \Sigma  = \Sigma^{gt}$ ); 
(2) informative with redundancy ($\hat\Sigma \supset \Sigma^{gt}$); 
(3) incomplete ($\hat \Sigma \subset \Sigma^{gt}$); 
(4) uninformative ($\hat\Sigma \not \subseteq \Sigma^{gt}$ and $\space \hat\Sigma \not \supseteq \Sigma^{gt}$).
Case (1) and (2) have ground-truth label informative (\texttt{Entailment}), while (3) and (4) are uninformative (\texttt{Not entailment}).

The evidence sets are created as follows.
For (2), we randomly sample distractors ($\Sigma^{gt} \cup d$) and add into the ground-truth set.
For (3), we randomly sample a strict subset from $\Sigma^{gt}$.
For (4), we repeatedly sample a set from all candidate evidence until the set satisfy the condition ($\hat\Sigma \not \subseteq \Sigma^{gt} \text{and} \space \hat\Sigma \not \supseteq \Sigma^{gt}$).

\paragraph{Verification methods}
We evaluate verification methods that output two-way classification labels: \{\texttt{Entailment}, \texttt{Not entailment}\}.
These two labels naturally exist in the multi-premise entailment benchmarks \cite{dalvi2021explaining, aghahadi2022avicenna, kamoi2023wice}. 

 We examine the characteristics of several verification methods, including 
(1) Natural language inference (NLI) models~\cite{he2021deberta};
(2) Large language models (LLMs), including GPT-4o~\cite{openai2024gpt4o}, Claude-3, Claude-3.5~\cite{anthropic2024claude}, and Llama-3.1 Instruct~\cite{meta2024llama3}.

\subsection{Results}

Classification results are presented in Table~\ref{tab:verification-prf1-results}.
We also present per-type accuracies in Table~\ref{tab:verification-results}.
We have the following observations. We also investigate the impact of different prompt structures on verification performance; a detailed analysis in Appendix~\ref{appendix:prompt_structures} shows that while more complex prompts can help in specific cases, they do not consistently outperform a basic, direct prompt.

\paragraph{NLI models are precise verifiers.}
It is found that NLI models achieve high precision across datasets (Table~\ref{tab:verification-prf1-results}).
However, they suffer from low recall on WiCE$_\text{claim}$ and WiCE$_\text{subclaim}$.
Further, NLI models are very conservative when predicting entailment (Table~\ref{tab:verification-results}). 

\paragraph{LLM verifiers tend to rationalize incomplete evidence with internal knowledge.}
From Table~\ref{tab:verification-results}, it can be observed that LLMs are prone to classify incomplete evidence sets (Inc.) as ``entailment'', leading to much worse than random performance on EntailmentBank.
This phenomenon is more pronounced in dataset with simpler languages (e.g., EntailmentBank). 
In contrast, supervised NLI classifiers are more conservative in terms of using internal knowledge. This behavior supports the notion that LLMs tend to fill gaps with their internal knowledge. This tendency to "fill in the gaps" highlights a significant reliability challenge. Future work could explore mitigation strategies, such as instruction-tuning models to explicitly forbid relying on internal knowledge, or incorporating more conservative verification mechanisms, like the NLI models we evaluated, to act as a safeguard against such rationalization.

\paragraph{Redundant evidence has little impact on model predictions.}
Though sensitive to uninformative or incomplete information, models are relatively robust to redundant information.

\begin{figure}
    \centering
    \includegraphics[width=\linewidth]{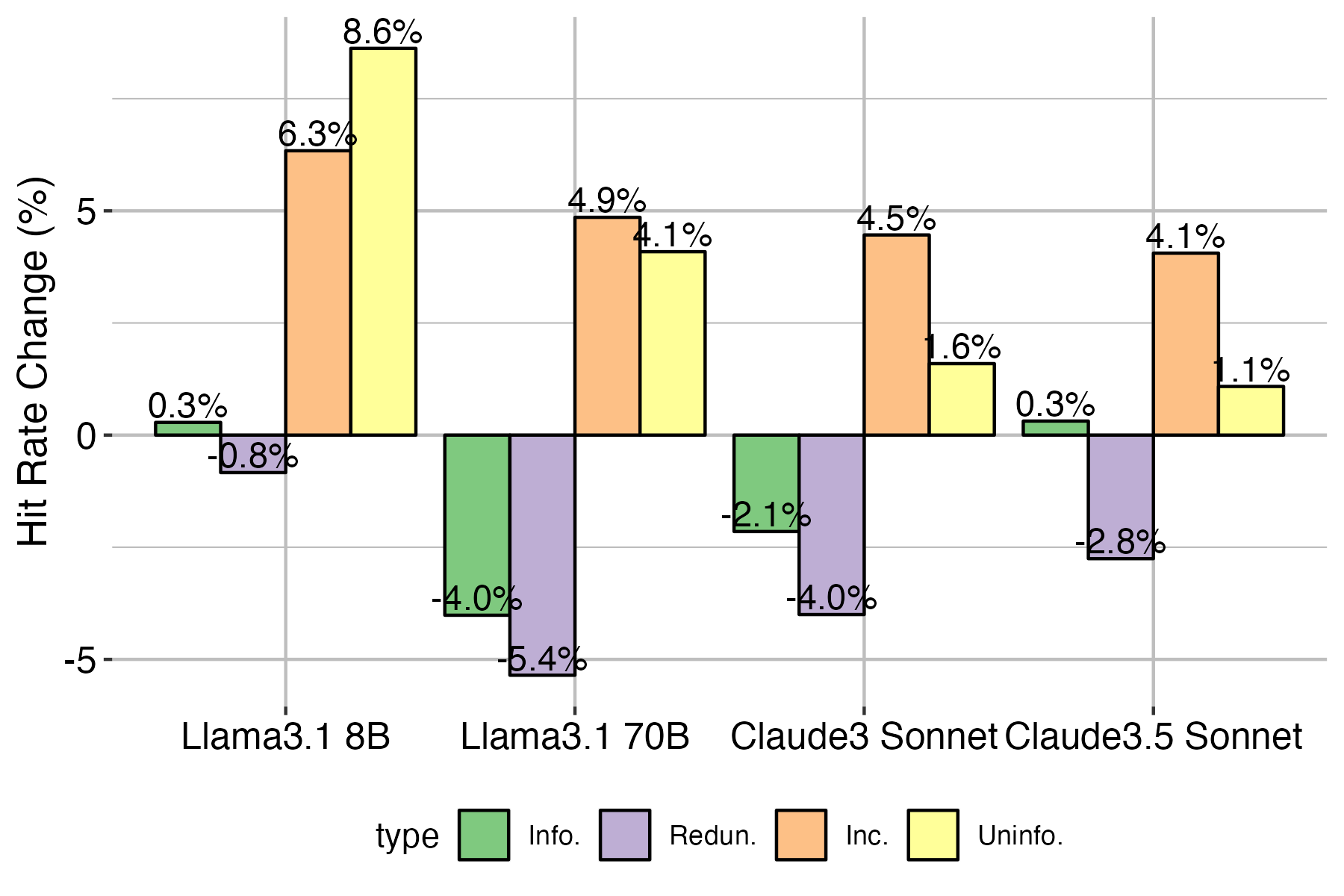}
    \caption{Hit Rate Change (\%) of LLMs + NLI model predictions. The numbers are averaged across four sources in \datasetname.}
    \label{fig:verif-ensemble}
\end{figure}

\paragraph{Combining LLM and NLI predictions leads to more conservative judgments.}
We investigate the effect of incorporating NLI predictions into LLM prompts on verification performance. As shown in Figure~\ref{fig:verif-ensemble}, this ensemble approach improves LLMs' ability to identify incomplete and uninformative instances, though at the cost of reduced accuracy in detecting informative and redundant evidence. This suggests that NLI models' more stringent criteria for entailment can help counteract LLMs' tendency to rationalize, albeit with a trade-off in overall verification performance.

\subsection{Qualitative analysis}

\begin{table}[ht!]
    \resizebox{0.5\textwidth}{!}{%
    \begin{tabular}{p{2cm}c}
        \toprule
        \multicolumn{1}{c}{\textbf{Hypothesis $\phi$} and \textbf{Evidences $\mathcal{K}$}} & \multicolumn{1}{c}{\textbf{Evaluation Sets $\hat \Sigma_t$} and \textbf{Predictions}}  \\
        \midrule
        
        \begin{tabular}[c]{@{}l@{}}\textbf{\textit{Hypothesis:}}\\ Northern hemisphere will have \\the most sunlight in summer.\\ \\ \textbf{\textit{Ground-truth Set:}}\\ {\color{purple}$k_1$}: The northern hemisphere is \\a kind of hemisphere of earth. \\ {\color{purple}$k_2$}: If a place is in summer, then \\it will have the most sunlight. \\ {\color{purple}$k_3$} A hemisphere of earth is a \\kind of place.\\ \\ \textbf{\textit{Sample distracter(s):}}\\ {\color{gray}$k_4$}: Daylight hours means time \\during which there is daylight.\\ {\color{gray}$k_5$}: Receiving sunlight is \\synonymous with absorbing \\sunlight.\\ {\color{gray}$k_6$} Period of daylight is \\synonymous with amount of \\daylight.\\ {\color{gray}$k_7$}: Sunshine means sunlight.\end{tabular} 
        
        &
        
        \begin{tabular}[c]{@{}l@{}}\textbf{\textit{Informative:}} {[}{\color{purple}$k_1$}, {\color{purple}$k_2$}, {\color{purple}$k_3$}{]} -- \textsc{ENT}\\ NLI-xlarge: \textsc{ENT}\\ Llama3.1 8B Instr.: \textsc{ENT}\\ Claude3 Sonnet: \textsc{ENT}\\ Claude3.5 Sonnet: \textsc{ENT}\\ \\ \textit{\textbf{Redundancy:}} {[}{\color{purple}$k_1$}, {\color{gray}$k_4$}, {\color{purple}$k_2$}, {\color{purple}$k_3$}{]} -- \textsc{ENT}\\ NLI-xlarge: \textsc{ENT}\\ Llama3.1 8B Instr.: \textsc{ENT}\\ Claude3 Sonnet: \textsc{ENT}\\ Claude3.5 Sonnet: \textsc{ENT}\\ \\ \textit{\textbf{Incomplete:}} {[}{\color{purple}$k_1$}{]} -- not \textsc{ENT}\\ NLI-xlarge: not \textsc{ENT}\\ Llama3.1 8B Instr.: \textsc{ENT}\\ Claude3 Sonnet: \textsc{ENT}\\ Claude3.5 Sonnet: not \textsc{ENT}\\ \\ \textit{\textbf{Uninformative:}} {[}{\color{gray}$k_5$}, {\color{gray}$k_6$}, {\color{gray}$k_7$}{]} -- not \textsc{ENT}\\ NLI-xlarge: not \textsc{ENT}\\ Llama3.1 8B Instr.: not \textsc{ENT}\\ Claude3 Sonnet: \textsc{ENT}\\ Claude3.5 Sonnet: not \textsc{ENT}\end{tabular} \\
        
        \bottomrule
    \end{tabular}
    }
    
    \caption{A case study of verification results. Ground-truth evidences are marked with {\color{purple}purple}, and distracting evidences are marked with {\color{gray}gray} for easier identification.
    ``\textsc{ENT}'' and `not \textsc{ENT}'' are short for dataset labels ``\texttt{Entailment}'' and ``\texttt{Not entailment}''.} 
    \label{tab:verification_examples}
\end{table}

A running example is presented in Table~\ref{tab:verification_examples}, showing how evaluation sets are constructed and how NLIs and LLMs perform. The NLI model, NLI-xlarge, shows consistent and precise verification capabilities across different evaluation sets. In contrast, LLMs like Llama3.1 8B and Claude models exhibit a tendency to rationalize incomplete evidence, as seen in the "Incomplete" set where they often incorrectly predict entailment.  The "Redundancy" set demonstrates that additional, non-essential information has minimal impact on model predictions.

\section{Retrieval planning for integrative grounding}
\label{sec:retrieval_planning}
In the sections above, we assume models are given a fixed set of queries (hypotheses) for retrieving evidence.
Recent advancements of proof systems \cite{sprague2022natural, tafjord2022entailer, weir2022dynamic} and LLM agents \cite{yao2022react, shinn2024reflexion} provide another perspective on this setting, where \textit{planning} is integrated to proactively intervene retrieval processes.
The objective of \textit{planning} is to increase the success rate of grounding, i.e., biasing the search space so that it is more likely for an informative set $\Sigma$ to be found.

The following part comprises experiments and discussions for three research questions: can planning and verification help retrieval and grounding? 
Moreover, how to optimize these components to maximize grounding performance?

\subsection{Evaluation setup}

\begin{table*}[ht!]
\centering
\resizebox{0.8\textwidth}{!}{
    \begin{tabular}{clccccc}
    \toprule
\multicolumn{1}{l}{}                                 &                           & \multicolumn{1}{c}{\textbf{No planning}} & \multicolumn{1}{c}{\textbf{Query Exp.}} & \multicolumn{1}{c}{\textbf{Fact Decomp.}} & \multicolumn{1}{c}{\textbf{Prop. Decomp.}} & \multicolumn{1}{c}{\textbf{Premise Abd.}} \\ \hline
\multicolumn{1}{c|}{\multirow{4}{*}{EntailmentBank}} & \multicolumn{1}{l|}{BM25} & 64.4                                     & 53.6                                    & \textbf{65.8}                             & {\ul 64.9}                                 & 56.3                                      \\
\multicolumn{1}{c|}{}                                & \multicolumn{1}{l|}{miLM} & {\ul 67.7}                               & 66.6                                    & 66.8                                      & 67.0                                       & \textbf{68.3}                             \\
\multicolumn{1}{c|}{}                                & \multicolumn{1}{l|}{mE5}  & {\ul 66.8}                               & 64.9                                    & 66.7                                      & 66.7                                       & \textbf{67.2}                             \\
\multicolumn{1}{c|}{}                                & \multicolumn{1}{l|}{ST5}  & {\ul 67.5}                               & 65.3                                    & 67.0                                      & 67.0                                       & \textbf{67.6}                             \\ \hline
\multicolumn{1}{c|}{\multirow{4}{*}{WiCE}}           & \multicolumn{1}{l|}{BM25} & \textbf{61.1}                            & 48.5                                    & 56.9                                      & {\ul 58.5}                                 & 52.4                                      \\
\multicolumn{1}{c|}{}                                & \multicolumn{1}{l|}{miLM} & \textbf{58.1}                            & 51.4                                    & {\ul 56.6}                                & {\ul 56.6}                                 & 53.9                                      \\
\multicolumn{1}{c|}{}                                & \multicolumn{1}{l|}{mE5}  & \textbf{64.0}                            & 58.8                                    & 61.5                                      & {\ul 62.6}                                 & 61.7                                      \\
\multicolumn{1}{c|}{}                                & \multicolumn{1}{l|}{ST5}  & \textbf{63.3}                            & 54.5                                    & 60.9                                      & {\ul 61.5}                                 & 59.1                                      \\ \hline
\multicolumn{1}{c|}{\multirow{4}{*}{HotpotQA}}       & \multicolumn{1}{l|}{BM25} & 67.5                                     & 65.8                                    & {\ul 68.7}                                & {\ul 68.7}                                 & \textbf{70.8}                             \\
\multicolumn{1}{c|}{}                                & \multicolumn{1}{l|}{miLM} & 69.5                                     & \textbf{73.9}                           & 68.0                                      & 67.8                                       & {\ul 72.3}                                \\
\multicolumn{1}{c|}{}                                & \multicolumn{1}{l|}{mE5}  & \textbf{80.1}                            & 79.3                                    & 75.2                                      & 75.8                                       & {\ul 79.7}                                \\
\multicolumn{1}{c|}{}                                & \multicolumn{1}{l|}{ST5}  & 71.7                                     & {\ul 71.8}                              & 71.7                                      & 71.7                                       & \textbf{74.0}                             \\ \hline
\multicolumn{1}{c|}{\multirow{4}{*}{MuSiQue}}        & \multicolumn{1}{l|}{BM25} & 60.9                                     & {\ul 64.1}                              & 57.8                                      & 58.4                                       & \textbf{67.7}                             \\
\multicolumn{1}{c|}{}                                & \multicolumn{1}{l|}{miLM} & 64.0                                     & \textbf{70.9}                           & 63.0                                      & 62.7                                       & {\ul 70.3}                                \\
\multicolumn{1}{c|}{}                                & \multicolumn{1}{l|}{mE5}  & 65.2                                     & \textbf{74.2}                           & 64.6                                      & 65.3                                       & {\ul 73.7}                                \\
\multicolumn{1}{c|}{}                                & \multicolumn{1}{l|}{ST5}  & 58.0                                     & {\ul 64.4}                              & 57.1                                      & 57.6                                       & \textbf{66.6}                             \\  \bottomrule
    \end{tabular}
}
\caption{Planning performance comparison based on Recall@5 (\%).  Best and second-best results are shown in \textbf{bold} and \underline{underlined}, respectively. Results represent mean performance values across different LLMs.}
\label{tab:planning-results}
\end{table*}

\paragraph{Retrievers}
For sparse retrievers, we evaluate BM25~\cite{robertson2009probabilistic}.
We also test dense retrievers, including MiniLM (miLM, \citealp{wang2020minilm}), Sentence T5 (ST5, \citealp{ni2021sentence}), and Microsoft E5-instruct (mE5, \citealp{wang2024multilingual}).
Given $\phi$ and $\mathcal{K}$, retrieval methods predict similarities and a ranking order of propositions in $\mathcal{K}$. 

\paragraph{Planning methods} 
Given the retrieval history $\Sigma_{t-1}$ and last step queries $\Phi_{t-1}$, we define ``planning'' as reasoning to generate a new set of queries $\Phi_t$ to guide the next retrieval step.
$$\Phi_t \leftarrow \texttt{Plan}(\Phi_{t-1}, \Sigma_{t-1})$$
Specifically, the planners we use can be summarized as follows.

Planners that do not depend on retrieval history ($\texttt{Plan}(\Phi_{t-1}, \varnothing)$), including:
\begin{itemize}[leftmargin=*]
    \item \textbf{Query expansion}~\cite{gao2023precise, wang2023query2doc}. This line of work expand writing based on the input query with an LLM.
    We adopt official prompts from HyDE~\cite{gao2023precise}.
    \item \textbf{Atomic fact decomposition}~\cite{min2023factscore, kamoi2023wice}. The hypothesis text is decomposed into multiple atomic factoids with a few-shot prompt. We reuse prompts from \cite{min2023factscore}.
    \item \textbf{Proposition decomposition}~\cite{chen2022propsegment}. Similar to atomic fact decomposition, proposition decomposition breaks down the input hypothesis text into multiple propositions. Since many have found that LLMs achieve reasonably good performance when prompted with few-shot examples \cite{min2023factscore, kamoi2023wice, chen2024sub}, we use the prompts provided in \cite{chen2024sub}.
    \item \textbf{Premise abduction}~\cite{tafjord2022entailer}. Given an input hypothesis text, premise abduction methods generate all premises required to entail the hypothesis through abduction. We curate few-shot prompts with examples from EntailmentBank~\cite{tafjord2022entailer}.
\end{itemize}


In addition, we also evaluate planners that take both input hypothesis and planning history as inputs ($\texttt{Plan}(\Phi_{t-1}, \Sigma_t)$).
This group of planners are closely related to the agentic behaviors of \textbf{self-reflection}~\cite{yao2022react, khot2022decomposed, shinn2024reflexion}.
Intuitively, integrative grounding may benefit from adjusting the queries to missing information and past queries. In our setup, this is implemented by prompting the LLM to analyze the retrieved evidence from the previous step, identify missing information, and generate a new, more targeted set of queries to guide the next retrieval iteration. The full prompt can be found in Appendix~\ref{appendix:prompts} (Table \ref{tab:prompts2}). Intuitively, integrative grounding may benefit from adjusting the queries to missing information and past queries.

We prompt three state-of-the-art LLMs, GPT-4o, Claude-3.5 Sonnet and Llama-3.1 70B Instruct models as the base LLMs for planning evaluation.
Each planning step produces multiple queries (in $\Phi_t$).
Following previous work in informative retrieval~\cite{gao2023precise, wang2023query2doc}, we concatenate them as the query for the next step retrieval. 
The implementation details are in Appendix~\ref{sec:appendix-experimental-details}.

\subsection{Results}

We compared the retrieval results of directly feeding hypotheses to retrievers ("No planning") against the performance of feeding both hypotheses and rewritten plans to retrievers. The main experimental results are presented in Table~\ref{tab:planning-results}.

\paragraph{Adding planning modules to refine the queries does not always help. In some cases, it can even hurt performance:} For instance, query expansion methods almost always led to decreased performance compared to no planning. This suggests that arbitrary query rewriting and expansion can introduce noise that hinders effective retrieval \cite{weller2023generative}.

\paragraph{Limited impact of decomposition-based planning:} Notably, planning based on atomic fact decomposition \cite{min2023factscore} or proposition decomposition \cite{chen2022propsegment} showed little improvement in grounding performance. We hypothesize that this is because decomposition-based planning does not introduce new information to the queries, potentially resulting in retrieved results that overlap significantly with the no-planning baseline.

\paragraph{Abduction-based planning shows significant improvement:} Among the four planning methods tested, Premise Abduction performed best. Although this method's intuition stems from strict textual entailment data \cite{dalvi2021explaining}, it appears to generalize well to broader datasets. This success could be attributed to the directed nature of such planning methods. Compared to "Query expansion" planning, premise abduction imposes an additional logical reasoning constraint (i.e., the possible premises of the hypothesis) to expand the search space effectively.

\begin{figure}[ht!]
    \centering
    \includegraphics[width=0.9\linewidth]{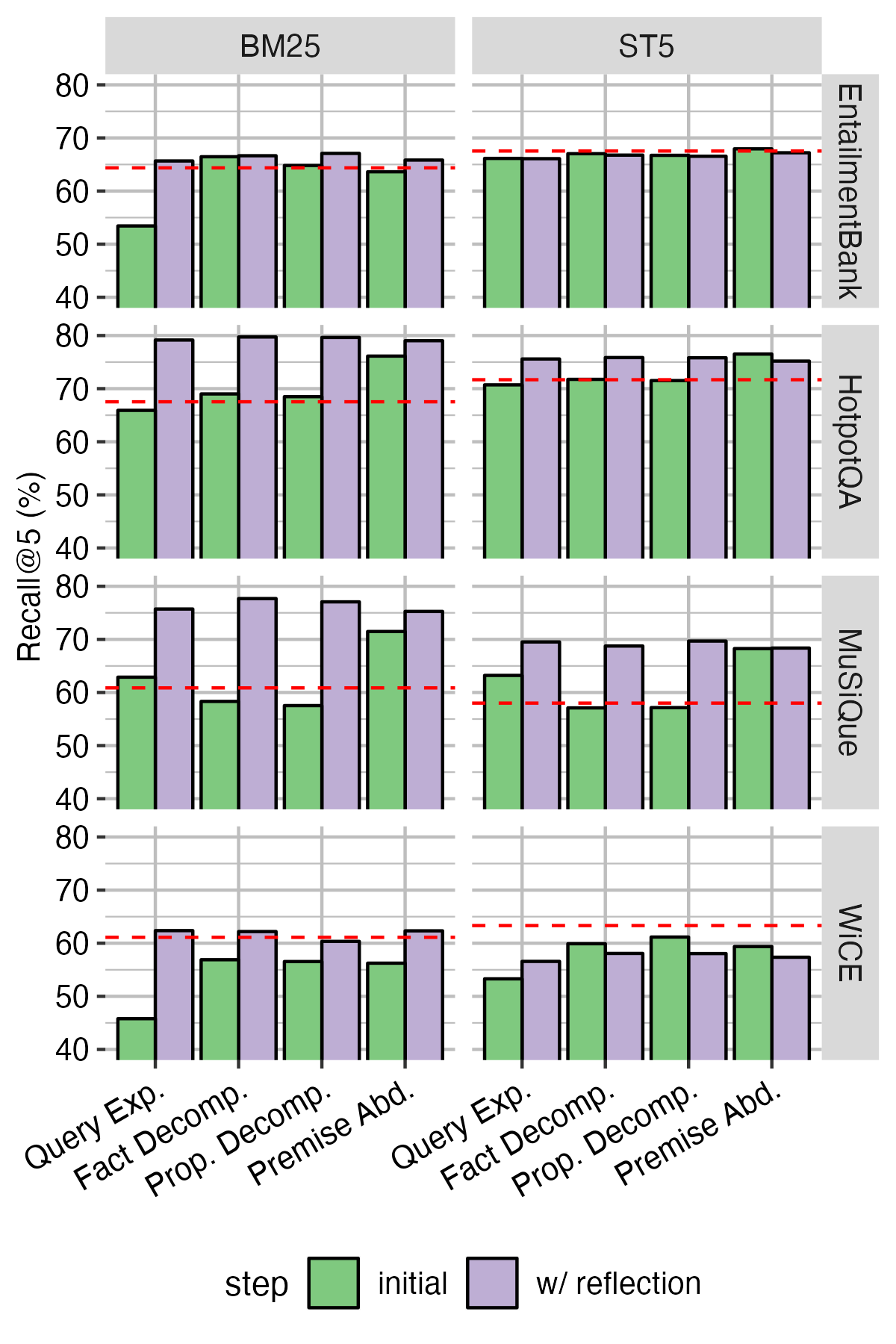}
    \caption{Performance comparison of planning with (``w/ reflection'') and without (``initial'') reflection  step using Recall@5 (\%). Dashed {\color{red}red} lines indicate baseline retrieval performance without planning.}
    \label{fig:planning}
    
\end{figure}

\paragraph{Self-reflection enhances integrative grounding:} As illustrated in Figure~\ref{fig:planning}, incorporating a zero-shot self-reflection step consistently improved the integrative grounding task. Most planning methods surpassed the No-planning baseline after the reflection step. Notably, query expansion and decomposition-based planning methods showed the greatest improvements. This may be because reflection helps mitigate the weaknesses of other planners: it provides a 'directed' bias that undirected query expansion lacks, and it introduces new information and context that conservative decomposition methods do not generate on their own.

\subsection{Qualitative analysis}
\begin{table}[ht!]
    \resizebox{0.5\textwidth}{!}{%
    \begin{tabular}{p{4cm}c}
        \toprule
        \multicolumn{1}{c}{\textbf{Planning}} & \multicolumn{1}{c}{\textbf{Retrieval results}}  \\
        \midrule
        \multicolumn{2}{l}{\textit{\textbf{Hypothesis}}: Northern hemisphere will have the most sunlight in summer.}\\
        \hline
                 
        \begin{tabular}[c]{@{}l@{}} \textbf{\textit{Proposition decomposition:}}\\ 
         \textbullet The northern hemisphere has \\more sunlight in summer. \\ \textbullet The northern hemisphere \\experiences seasons.\\ \textbullet Summer is a season in the \\northern hemisphere.\\ \textbullet Sunlight varies by season in \\the northern hemisphere.\\ \end{tabular}
        &
        \begin{tabular}[c]{@{}l@{}} {\color{purple}\textbullet The northern hemisphere is a} \\{\color{purple}kind of hemisphere of earth.}\\ {\color{purple} \textbullet If a place is in summer, then it}\\ {\color{purple}will have the most sunlight.}\\ \textbullet A hemisphere is a part of earth. \\ \textbullet Being in the sun is synonymous \\with being in the sunlight.\\  {\color{purple} \textbullet A hemisphere of earth is a kind}\\ {\color{purple}of place.} \end{tabular} \\
        \hline 

        \begin{tabular}[c]{@{}l@{}} \textit{\textbf{Atomic fact decomposition:}}\\ \textbullet The northern hemisphere \\experiences seasons.\\ \textbullet Summer is one of the seasons.\\ \textbullet The northern hemisphere \\receives sunlight.\\ \textbullet The northern hemisphere \\receives the most sunlight in \\summer.\end{tabular}\
        &
        \begin{tabular}[c]{@{}l@{}} {\color{purple}\textbullet The northern hemisphere is a kind} \\ {\color{purple}of hemisphere of earth.}\\ {\color{purple} \textbullet If a place is in summer, then it}\\ {\color{purple}will have the most sunlight.}\\ \textbullet Being in the sun is synonymous \\with being in the sunlight.\\ \textbullet A hemisphere is a part of earth.\\  {\color{purple} \textbullet A hemisphere of earth is a kind}\\ {\color{purple}of place.} \\ \end{tabular}\\

        \bottomrule
    \end{tabular}
    }
    \caption{Illustration of the planning process. The "Planning" column shows representative queries generated by Claude-v3.5-sonnet. For brevity, only proposition decomposition and atomic fact decomposition are presented. The "Retrieval" column displays evidence pieces retrieved using BM25. Ground truth statements are highlighted in {\color{purple}purple}.} 
    \label{tab:planning_examples}
\end{table}
The example in Table \ref{tab:planning_examples} illustrates the workings of different planning methods. The proposition decomposition method breaks down the hypothesis into four key components, focusing on the relationship between the northern hemisphere, seasons, and sunlight variation. Similarly, the atomic fact decomposition method generates four fundamental statements about these elements. Notably, neither method introduces new information beyond the original hypothesis. In this case, both methods capture all three ground truth evidences.


\section{Related work}
\noindent \textbf{Natural Language Proof Generation.} Prior work has developed proof writing algorithms that generate proof trees based on models' internal knowledge~\cite{tafjord2022entailer, sprague2022natural}. While NELLIE~\cite{weir2022dynamic} incorporates retrieved facts for hypothesis decomposition, these approaches typically operate in restricted domains. More recent work has also focused on enhancing this process using principles from informal logic to improve decompositional inference~\cite{weir2024enhancing}.

\noindent \textbf{Fact Verification with LLMs.} Recent work on fact-checking LLM-generated content~\cite{min2023factscore, tang2024minicheck, rashkin2023measuring} primarily focuses on single-premise verification. Other studies have investigated how well LLMs ground their outputs in provided sources, confirming that even state-of-the-art models struggle with faithfully adhering to evidence~\cite{lee2023well}, especially when it contains conflicting information~\cite{jiayang2024econ}. While related to our verification component, our work focuses on a comprehensive evaluation of integrative retrieval and planning.


\noindent \textbf{RAG and Multi-hop QA Agents.} Recent LLM agents for multi-hop question answering~\cite{yao2022react, shinn2024reflexion} employ iterative retrieval strategies but focus primarily on reasoning rather than addressing integrative grounding challenges. Methods like TRACE~\cite{fang2024trace} construct reasoning chains from already-retrieved evidence, whereas our approach emphasizes dynamically planning what to retrieve next based on evidence interdependencies. Similarly, while studies like~\cite{trautmann2024measuring} focus on post-generation evaluation and others~\cite{song2024measuring} enhance citation quality, we address verification during the retrieval process to evaluate whether multiple documents collectively support a hypothesis.

\section{Conclusion}

In this work, we introduce "integrative grounding" as a critical challenge for LLMs and provide a systematic evaluation framework, InteGround, to assess it. Our investigation yields several key insights with direct implications for building more reliable systems. We demonstrate that while LLMs are robust to redundant evidence, they exhibit a strong tendency to "rationalize" with internal knowledge when faced with incomplete information, posing a significant risk to faithfulness. In retrieval planning, we find that intuitive strategies like undirected query expansion can degrade performance by introducing noise, whereas logically constrained methods show significant promise. Notably, premise abduction prove effective by expanding the search space in a directed manner, and zero-shot self-reflection consistently improve performance across all planning methods by enabling iterative refinement.
Our findings offer direct guidance for building more robust grounding systems, with a detailed discussion of practical applications for RAG pipelines provided in Appendix~\ref{appendix:rag_system_discussion}.


\section*{Limitations}

\noindent \textit{Limited ground-truth setting.} This study assumes that a single, unique ground-truth evidence set. However, in many real-world scenarios, multiple valid ground-truth evidence sets may exist to support a hypothesis. 

\noindent \textit{Grounding to structured data.} Our evaluation is restricted to the textual domain. Grounding to structured or semi-structured data, such as tabular data or knowledge graphs, is also an important and promising direction for future research. 

\noindent \textit{Limited language.} We primarily focus on English corpora, meaning our findings may not generalize to other languages. Evaluation of grounding on multilingual corpora is left for future work.

\noindent \textit{Evaluation-centric approach.} Our work prioritizes the comprehensive evaluation of integrative grounding rather than direct application development. While our findings have implications for RAG systems, translating these insights into optimized real-world applications requires additional engineering effort beyond the scope of this study.

\noindent \textit{Limited Scope of Planners.} Our study focuses exclusively on LLM-driven planning strategies. A comparison with established non-LLM planning techniques, such as classical symbolic planners, was beyond our scope but remains an important direction for future comparative analysis.

\section*{Ethical statement}
Our work primarily utilizes open-source retrieval models, datasets, and publicly available Large Language Models (LLMs). Given the nature of our research context, the outputs generated by these LLMs are unlikely to contain harmful or dangerous information. We have carefully considered the ethical implications of our study and foresee no significant concerns or potential risks associated with our methodology or findings. 

\section*{Acknowledgements}
The authors of this paper were supported by the ITSP Platform Research Project (ITS/189/23FP) from ITC of Hong Kong, SAR, China, and the AoE (AoE/E-601/24-N), the RIF (R6021-20) and the GRF (16205322) from RGC of Hong Kong, SAR, China. 
We would like to express our sincere gratitude to all the reviewers for their invaluable contributions through their comments and suggestions.

\bibliography{custom}
\bibliographystyle{acl_natbib}

\appendix

\section{Appendix}
\label{sec:appendix}

\subsection{Details in the construction of \datasetname}
\label{sec:appendix-data-construction}

\begin{table}
    \centering
\resizebox{0.9\linewidth}{!}{
\begin{tabular}{l|lll}
\toprule
               & \multicolumn{1}{c}{$\phi$} & \multicolumn{1}{c}{$\Sigma^{gt}$} & \multicolumn{1}{c}{\cellcolor[HTML]{FFFFFF}$\mathcal{K}$} \\ \hline
EntailmentBank & 11.8±4.3                   & 9.5±4.3                           & 9.4±5.0                                                   \\ \hline
WiCE           & 25.9±11.7                  & 21.7±16.2                         & 12.5±13.8                                                 \\ \hline
HotpotQA       & 21.4±7.4                   & 34.4±13.7                         & 33.1±13.3                                                 \\ \hline
MuSiQue        & 25.9±6.3                   & 110.2±63.9                        & 105.2±59.0                                                \\ \bottomrule
\end{tabular}
    }
    \caption{Mean and standard deviation for numbers of tokens in \datasetname.}
    \label{tab:data-statistics-additional}
\end{table}
The number of tokens are presented in Table~\ref{tab:data-statistics-additional}.

\subsection{Experimental details}
\label{sec:appendix-experimental-details}

\subsubsection{Baselines}
Retrieval methods:
\begin{itemize}
    \item BM25 \cite{robertson2009probabilistic}. We use the \texttt{rank-bm25} python package to implement the algorithm.
    \item Sentence-transformers. We use the LangChain\cite{topsakal2023creating} implementation to embed corpus, and the cosine similarities between embeddings as the similarity for retrieval.
\end{itemize}

Verification and planning methods:
\begin{itemize}
    \item NLI. We use the state-of-the-art NLI models \cite{he2020deberta}, including DeBERTa (\texttt{xlarge}) and DeBERTa-v2 (\texttt{xxlarge}). Given a pair of texts,  NLI models output probabilities over entailment, contradiction, and neutral (\textsc{ENT}, \textsc{CON}, \textsc{NEU}).
    \item LLMs. We use the state-of-the-art LLMs, including Llama 3.1 8B Instruct, Llama 3.1 70B Instruct~\cite{llama3}, Claude 3 Haiku, Claude 3 Sonnet, and Claude 3.5 Sonnet~\cite{claude3}. Llama and Claude models are accessed through Amazon Bedrock.
\end{itemize}

\subsubsection{Experimental settings}
\noindent \textbf{Direct retrieval}
Ranking evaluation. We use the \texttt{ranx}~\cite{ranx} \footnote{https://github.com/AmenRa/ranx} package for computing ranking metrics in the evaluation of retrievers. In the main evaluation, F1@5 and Acc@5 are reported.

\begin{table*}
    \centering
    \resizebox{0.9\textwidth}{!}{
\begin{tabular}{l|cc|cc|cc|cc}
\toprule
                               & \multicolumn{2}{c|}{EntailmentBank} & \multicolumn{2}{c|}{WiCE}     & \multicolumn{2}{c|}{HotpotQA} & \multicolumn{2}{c}{MuSiQue}   \\ \hline
                               & F1@5             & Acc@5            & F1@5          & Acc@5         & F1@5          & Acc@5         & F1@5          & Acc@5         \\ \hline
BM25                           & 52.5             & 34.1             & 40.7          & 30.5          & 53.3          & 26.6          & 48.4          & 15.0          \\
SimCSE$_{\text{RoBERTa}}$ & 50.2             & 29.7             & 38.8          & 25.6          & 50.4          & 21.8          & 40.6          & 8.6           \\
MiniLM-L6                      & 55.3             & 36.2             & 38.5          & 25.6          & 55.1          & 27.8          & 50.9          & 17.6          \\
ST5$_{\text{large}}$           & 55.1             & 37.1             & \textbf{42.0} & 31.2          & 56.7          & 32.8          & 46.1          & 14.2          \\
GTR$_{\text{T5-large}}$        & \textbf{57.6}    & \textbf{40.0}    & 41.3          & \textbf{32.3} & 56.6          & 32.8          & 50.9          & 18.8          \\
mE5$_{\text{large-instruct}}$  & 54.5             & 37.1             & 41.2          & 30.2          & \textbf{62.8} & \textbf{48.0} & \textbf{54.4} & \textbf{24.2} \\\bottomrule
        \end{tabular}
    }
    \caption{Direct retrieval results.}
    \label{tab:direct_retrieval}
\end{table*}

\noindent \textbf{Stepwise retrieval}
We add ground-truth evidence step by step to the hypothesis.

\paragraph{Combining planning and retrieval} 
In this setting, we generate multiple rankings by retrieving with different sub-queries. To consolidate these rankings into a single ranking, we address it as a rank aggregation problem ~\cite{dwork2001rank}. We implement Borda's rank aggregation strategy~\cite{borda1781m} to produce a unified rank.

\subsubsection{LMs prompting details}
\label{appendix:prompts}
The prompt templates for LLMs in this research are presented in Table \ref{tab:prompts1} for hypothesis generation and verifications and Table \ref{tab:prompts2} for plannings.
\begin{table*}[t]
\resizebox{\textwidth}{!}{%
    \centering
    \small
    \begin{tabular}{p{0.1\textwidth}p{0.2\textwidth}p{0.7\textwidth}}
    \toprule
         \multicolumn{1}{c}{\textbf{Function}} & \multicolumn{1}{c}{\textbf{Inputs}} & \multicolumn{1}{c}{\textbf{Prompt}}  \\
    \midrule

    Hypothesis Generation
    &
    $q$: question
    
    $a$: answer 
    & 
    Paraphrase the given question and answer pair to a proposition. Your response should be formatted as \{\{"Proposition": "PROPOSITION TEXT"\}\}.
    
    {\color{white}blank}
    
    Question: When did the maker of the Acura Legend, the manufacturer of Toyopet Master, and Nissan open US assembly plants?

    Answer: 1981

    \{\{"Proposition": "The maker of the Acura Legend, the manufacturer of Toyopet Master, and Nissan opened US assembly plants in 1981."\}\}   

    {\color{white}blank}
    
    Question: Signed with Maybach Music Group in 2011, which artist was featured as a guest in Fire of Zamani?
    
    Answer: Wale
    
    \{\{"Proposition": "Wale, who signed with Maybach Music Group in 2011, was a featured guest artist on Fire of Zamani."\}\}

    {\color{white}blank}
       
    Question: \{$q$\}
    
    Answer: \{$a$\} \\
    \hline

    Verification (LLMs)
    &
    $e_1$: evidence set   
    
    $e_2$: hypothesis  
    &
    You are a helpful logical reasoner. Please help classify a hypothesis with \{labels\} based solely on a set of evidence.

    {\color{white}blank}
    
    Evidence set:\{$e_1$\}
    
    Hypothesis: \{$e_2$\}
    
    Result in JSON format (e.g. \{\{"label": "\{labels\}"\}\}):\\
    \hline

    Verification (NLIs+LLMs)
    &
    $label$: NLI's prediction
    
    $e_1$: evidence set   
    
    $e_2$: hypothesis  
    &
    You are a helpful logical reasoner. Please help classify a hypothesis with \{labels\} based solely on a set of evidence.

    {\color{white}blank}
    
    For your reference, an external supervised Natural Language Inference model's prediction is: \{$label$\}.
    
    {\color{white}blank}
    
    Evidence set: \{$e_1$\}
    
    Hypothesis: \{$e_2$\}
    
    Result in JSON format (e.g. \{\{"label": "\{labels\}"\}\}):\\

    \bottomrule
    \end{tabular}
}
    \caption{Prompts for LLMs: Hypothesis Generation and Verifications}
    \label{tab:prompts1}

\end{table*}

\begin{table*}[t]
\resizebox{\textwidth}{!}{%
    \centering
    \small
    \begin{tabular}{p{0.1\textwidth}p{0.15\textwidth}p{0.9\textwidth}}
    \toprule
         \multicolumn{1}{c}{\textbf{Function}} & \multicolumn{1}{c}{\textbf{Inputs}} & \multicolumn{1}{c}{\textbf{Prompt}}  \\
    \midrule
    Planning 
    
    (Premise abduction)
    &
    $k$: hypothesis
    &
    Given the following hypothesis, try to generate a set of premises that can prove the hypothesis. Please format the premises as \{\{"Premises": ["PREMISE 1 TEXT", "PREMISE 2 TEXT", ...]\}\}.

    {\color{white}blank}

    Hypothesis: The earth revolving around the sun causes leo to appear in different areas in the sky at different times of year.

    \{\{"Premises": ["Leo is a kind of constellation.", "A constellation contains stars.", "The earth revolving around the sun causes stars to appear in different areas in the sky at different times of year."]\}\}

    {\color{white}blank}

    Hypothesis: The earth rotating on its axis causes stars to move relative to the horizon during the night.
    
    \{\{"Premises": ["Apparent motion is when an object appears to move relative to another object's position.", "The earth rotating on its axis causes stars to appear to move across the sky at night.", "Earth is a kind of celestial object.", "A star is a kind of celestial object / celestial body.", "Stars appear to move relative to the horizon during the night."]\}\}

    {\color{white}blank}

    Hypothesis: \{$k$\} \\
    \hline

    Planning 
    
    (Atomic fact decomposition)
    &
    $s$: sentence
    &
    Example 0:
    
    Please breakdown the following sentence into independent facts: He made his acting debut in the film The Moon is the Sun’s Dream (1992), and continued to appear in small and supporting roles throughout the 1990s.
    
    \{\{"facts": ["He made his acting debut in the film.", "He made his acting debut in The Moon is the Sun's Dream.", "The Moon is the Sun's Dream is a film.", "The Moon is the Sun's Dream was released in 1992.", "After his acting debut, he appeared in small and supporting roles.", "After his acting debut, he appeared in small and supporting roles throughout the 1990s."]\}\}

    {\color{white}blank}
    
    Example 1:
    
    Please breakdown the following sentence into independent facts: He is also a successful producer and engineer, having worked with a wide variety of artists, including Willie Nelson, Tim McGraw, and Taylor Swift.

    \{\{"facts": ["He is successful.", "He is a producer.", "He is a engineer.", "He has worked with a wide variety of artists.", "Willie Nelson is an artist.", "He has worked with Willie Nelson.", "Tim McGraw is an artist.", "He has worked with Tim McGraw.", "Taylor Swift is an artist.", "He has worked with Taylor Swift."]\}\}

    {\color{white}blank}
    
    Example 2:
    
    Please breakdown the following sentence into independent facts: In 1963, Collins became one of the third group of astronauts selected by NASA and he served as the back-up Command Module Pilot for the Gemini 7 mission.
    
    \{\{"facts": ["Collins became an astronaut.", "Collins became one of the third group of astronauts.", "Collins became one of the third group of astronauts selected.", "Collins became one of the third group of astronauts selected by NASA.", "Collins became one of the third group of astronauts selected by NASA in 1963.", "He served as the Command Module Pilot.", "He served as the back-up Command Module Pilot.", "He served as the Command Module Pilot for the Gemini 7 mission."]\}\}

    {\color{white}blank}
    
    Example 3:
    
    Please breakdown the following sentence into independent facts: \{$s$\}\\
    \hline 

    Planning 
    
    (Proposition decomposition)
    &
    $s$: sentence
    &
    Given the following sentence, tell me what claims they are making. Please split the sentence as much as possible, but do not include information not in the sentence.

    {\color{white}blank}
    
    Sentence: The Andy Warhol Museum in his hometown, Pittsburgh, Pennsylvania, contains an extensive permanent collection of art.
    
    \{\{"Claims": ["The Andy Warhol Museum is in Pittsburgh.", "Andy Warhol’s hometown is in Pittsburgh.", "Pittsburgh is in Pennsylvania.", "The Andy Warhol Museum contains an extensive permanent collection of art."]\}\}

    {\color{white}blank}
    
    Sentence: \{$s$\} \\   
    \hline 

    Planning 
    
    (Query expansion)
    &
    $k$: claim
    &
    Please write a passage to support/refute the claim.
   
    Claim: {$k$}

    Passage (in the format "\{\{"passage": "PASSAGE TEXT"\}\}"):\\
    \hline 
    Planning 
    
    (with history)
    &
    $k$: hypothesis

    $q$: previous queries
    
    $s$: previous search results 
    &
    You are an AI information retrieval specialist trained to optimize search queries for finding relevant evidence in factual sources.

    {\color{white}blank}

    Task: Generate targeted search queries to find evidence that could either support or disprove the given hypothesis.
    
    {\color{white}blank}

    Requirements:
    
    1. Generate 3-5 refined search queries
    
    2. Each query should be specific and focused
    
    3. Consider both supporting and contradicting evidence
    
    4. You may retain effective queries from the previous round

    {\color{white}blank}
    
    Input Hypothesis: {$k$}

    {\color{white}blank}

    Previous Information:
    
    - Previous queries: {$q$}
    
    - Previous search results: {$s$}

    {\color{white}blank}

    Output Format:
    
    \{\{"queries": ["QUERY TEXT 1", "QUERY TEXT 2", ...]\}\}\\
        
    \bottomrule
    \end{tabular}
}
    \caption{Prompts for LLMs: Planning}
    \label{tab:prompts2}

\end{table*}

\subsubsection{Additional experiment on prompt structures}
\label{appendix:prompt_structures}
\begin{table*}[ht!]
\centering
\resizebox{0.9\linewidth}{!}{
\begin{tabular}{l|rrr|rrr|rrr|rrr}
\toprule
           & \multicolumn{3}{c|}{EntailmentBank}                                     & \multicolumn{3}{c|}{WiCE}                                               & \multicolumn{3}{c|}{HotpotQA}                                           & \multicolumn{3}{c}{MuSiQue}                                            \\
           & \multicolumn{1}{c}{P} & \multicolumn{1}{c}{R} & \multicolumn{1}{c|}{F1} & \multicolumn{1}{c}{P} & \multicolumn{1}{c}{R} & \multicolumn{1}{c|}{F1} & \multicolumn{1}{c}{P} & \multicolumn{1}{c}{R} & \multicolumn{1}{c|}{F1} & \multicolumn{1}{c}{P} & \multicolumn{1}{c}{R} & \multicolumn{1}{c}{F1} \\ \hline
Basic      & 63.6                  & 97.2                  & 76.9                    & 88.1                  & 66.0                  & 75.4                    & 83.5                  & 83.8                  & 83.7                    & 82.9                  & 49.8                  & 62.2                   \\ \hline
Structured & 77.2                  & 86.0                  & 81.4                    & 96.4                  & 47.4                  & 63.5                    & 90.0                  & 79.6                  & 84.5                    & 88.9                  & 44.1                  & 59.0                   \\ \hline
CoT        & 73.1                  & 88.5                  & 80.1                    & 90.6                  & 52.5                  & 66.4                    & 85.1                  & 79.9                  & 82.4                    & 86.0                  & 50.2                  & 63.4                   \\ \bottomrule
\end{tabular}
}
\caption{Comparison of three different prompting schemes in groundedness verification.} 
    \label{tab:verification_prompt_type}

\end{table*}

\begin{table*}[ht!]
\centering
\resizebox{0.9\linewidth}{!}{
\begin{tabular}{l|rrrr|rrrr|rrrr|rrrr}
\hline
\multirow{2}{*}{} & \multicolumn{4}{c|}{\textbf{EntailmentBank}}                                                                     & \multicolumn{4}{c|}{\textbf{WiCE}}                                                                               & \multicolumn{4}{c|}{\textbf{HotpotQA}}                                                                           & \multicolumn{4}{c}{\textbf{MuSiQue}}                                                                            \\
                  & \multicolumn{1}{c}{Info.} & \multicolumn{1}{c}{Redun.} & \multicolumn{1}{c}{Inc.} & \multicolumn{1}{c|}{Uninfo.} & \multicolumn{1}{c}{Info.} & \multicolumn{1}{c}{Redun.} & \multicolumn{1}{c}{Inc.} & \multicolumn{1}{c|}{Uninfo.} & \multicolumn{1}{c}{Info.} & \multicolumn{1}{c}{Redun.} & \multicolumn{1}{c}{Inc.} & \multicolumn{1}{c|}{Uninfo.} & \multicolumn{1}{c}{Info.} & \multicolumn{1}{c}{Redun.} & \multicolumn{1}{c}{Inc.} & \multicolumn{1}{c}{Uninfo.} \\ \hline
Basic             & 97.4                      & 97.1                       & 34.4                     & 54.4                         & 68.4                      & 63.5                       & 85.3                     & 96.8                         & 82.2                      & 85.4                       & 74.8                     & 92.2                         & 48                        & 51.6                       & 84.2                     & 95.2                        \\ \hline
Structured        & 88.2                      & 83.8                       & 62.9                     & 86.2                         & 49.8                      & 44.9                       & 96.8                     & 99.6                         & 79.4                      & 79.8                       & 87.4                     & 95                           & 44.6                      & 43.6                       & 90.8                     & 98.2                        \\ \hline
CoT               & 88.8                      & 88.2                       & 57.4                     & 77.4                         & 53                        & 51.9                       & 90.9                     & 98.2                         & 78.4                      & 81.4                       & 79.6                     & 92.4                         & 51.4                      & 49                         & 88                       & 95.6                        \\ \bottomrule
\end{tabular}
}
\caption{Per-type performance in the comparison of three different prompting schemes in groundedness verification.} 
    \label{tab:verification_prompt_type_2}

\end{table*}
As shown in Table~\ref{tab:verification_prompt_type}, we test GPT-4o models with different prompts on the groundedness verification task.

\noindent \textbf{Basic Prompt} (in the paper): Simple instruction to assess if evidence supports a hypothesis query.

``You are a helpful logical reasoner. Please help classify a hypothesis with \{labels\} based solely on a set of evidence. Evidence set: \{e1\} Hypothesis: \{e2\}''

\noindent \textbf{Structured Reasoning Prompt}: Include explicit steps for verification (check completeness, redundancy, etc.)

``Evidence set: \{e1\} Hypothesis: \{{e2\} Assess whether the evidence is sufficient to support the query by checking: 1. Relevance: Is all the evidence relevant to the query? 2. Completeness: Does the evidence contain all necessary information to address the query? 3. Redundancy: Is there unnecessary repetition in the evidence? Based on this assessment, is the provided evidence sufficient to support the hypothesis? Briefly explain your assessment, then choose your answer among \{labels\}.''

\noindent \textbf{Chain-of-Thought Prompt}: Ask the model to think step-by-step before concluding

``Evidence set: \{e1\} Hypothesis: \{e2\} Think step by step to determine if the provided evidence is sufficient to support the hypothesis. The following steps are an example: - What key information does the query require? - What information does the evidence provide? - What information, if any, is missing? - What additional evidence would be needed to fully address the query? After thinking step by step, determine if the provided evidence is sufficient to support the hypothesis. Choose your answer among \{labels\}.''

We have the following observations:

\begin{itemize}
    \item Prompts with complex structures (Structured, CoT) enhance models' detection of incomplete information, but often reduce accuracy on "Informative" labels—suggesting a potential overthinking effect.
    \item Despite this trade-off, overall classification performance (F1) remains robust across prompting schemes, with complex prompts did not outperform the Basic prompting, and in some domains, performed significantly worse."
\end{itemize}

\subsubsection{Applications to RAG Systems}
\label{appendix:rag_system_discussion}
Our research offers significant practical applications for enhancing RAG systems. First, our analysis of planning strategies provides actionable methods for improving evidence retrieval in production pipelines. Specifically, premise abduction addresses cases with incomplete evidence by generating plausible intermediate premises that guide subsequent retrievals. Similarly, fact decomposition simplifies complex queries, substantially improving retrieval accuracy in noisy information environments.

Second, our groundedness verification findings directly inform RAG system design. The observed tendency of LLMs to rationalize when evidence is incomplete underscores the need for dedicated verification mechanisms to detect and mitigate hallucinations. Our evaluation framework offers a approach for evaluating the effectiveness of such safeguards in practical applications.

Finally, the performance disparities across different verification strategies provide clear guidance for RAG system architecture decisions. By incorporating these insights, developers can create more reliable systems that not only retrieve relevant information but also accurately assess whether the retrieved evidence collectively supports the generated content.

}

\subsection{Grounding systems }
\label{sec:def-system}
A grounding system serves the goal of finding a subset $\Sigma$ from a KB $\mathcal{K}$, given a hypothesis $\phi$.
Although there may be various ways to achieve this goal, there are common stages among all the grounding systems: 
the \textit{planning} stage which involves reasoning over the hypothesis, and 
the \textit{linking} stage which retrieves candidates from $\mathcal{K}$ and verifies the groundedness of such candidates set.
In the literature of logical reasoning \cite{poole2010artificial}, forward chaining and backward chaining provides insights on the possible implementations of the stages.

Suppose each grounding has at most $T$ ($T\geq 1$) steps.
Let $\Sigma_{t}$ denote the grounded set and $\Phi_t$ denote the hypotheses tree at time step  $t$ ($t\in\{0, 1, \cdots, T\}$).
Initially,  $\Sigma_{0}=\{\}$ and $\Phi_0$ contains only the root node $\phi$. 

\noindent \textit{\textbf{Linking.}} 
The system first conduct linking to update the candidate set:
$$\hat\Sigma_t \leftarrow \texttt{Retrieve}(\Sigma_{t-1}, \Phi_{t-1}, \mathcal{K})$$
The candidate set is then judged for testing whether the hypotheses are consistent with, using the \texttt{Ask} function, where the qualifies subset is retained
$$\Sigma_t \leftarrow \texttt{Verify}(\texttt{Ask}_{\hat\Sigma_t}(\Phi_{t-1}))$$
The linking process trigger exiting condition when \texttt{Ask}$_{\Sigma_t}(\Phi_{t-1})$ returns \textit{informative} response for all the leaf nodes in $\Phi_{t-1}$. 

Essentially, forward chaining is applied to test whether the hypothesis follows $\Sigma_t$.

\noindent \textit{\textbf{Planning.}} 
Given $\Sigma_{t}$ and $\Phi_{t-1}$, a grounding system do reasoning to update the hypotheses tree so as to guide the next linking step.
$$\Phi_t \leftarrow \texttt{Plan}(\Sigma_{t}, \Phi_{t-1})$$

Although how to implement the reasoning function here is up to each grounding system's design,
this reasoning stage is essentially backward-chaining. Developing robust reasoning functions is a significant challenge, as LLMs often fall short on complex cognitive tasks like story-level analogy~\cite{jiayang2023storyanalogy,zheng2025logidynamics}. Backward chaining, or abductive reasoning based on tree $\Phi_{t-1}$ and premises set $\Sigma_t$, can provide additional coverage for searching over $\mathcal{K}$ \cite{sprague2022natural, tafjord2022entailer}.

\end{document}